\documentclass[a4paper,11pt,twocolumn,twoside]{article}
\usepackage[dvips]{graphicx}
\usepackage{sepln_en}
\usepackage{fullname}
\usepackage[utf8]{inputenc}
\usepackage[english]{babel}
\usepackage{libertine}
\usepackage{numprint} 
\usepackage{graphicx}
\usepackage{floatflt}
\usepackage{blindtext}
\usepackage{enumitem}
\usepackage{amsthm}
\usepackage{subfig}
\usepackage{listings}
\usepackage{listingsutf8}
\usepackage[breaklinks]{hyperref}
\usepackage{amsmath}
\usepackage{framed}
\usepackage{minibox}
\usepackage{comment}
\usepackage{float}
\usepackage{wrapfig}
\usepackage{longtable}
\usepackage[strict]{changepage}
\usepackage{pgfplots}
\usepackage{tikz}
\usepackage{booktabs}
\usepackage{multirow}
\usepackage{rotating}

\usepackage{amssymb}
\usepackage{marvosym}
\usepackage{makecell}
\usepackage[acronym]{glossaries}
\makeglossaries

\newacronym{LLM}{LLM}{Large Language Models}
\newacronym{TS}{TS}{Text Simplification}
\newacronym{FEINA}{FEINA}{Financial Education Corpus IN SpAnish}
\newacronym{AI}{AI}{Artificial Intelligence}
\newacronym{BERT}{BERT}{Bidirectional Encoder Representations from
Transformers}

\newacronym{BETO}{BETO}{Spanish BERT}

\newacronym{NLP}{NLP}{Natural Language Processing}

\newacronym{UCCA}{UCCA}{Universal Cognitive Conceptual
Annotation}
\newacronym{EASSE}{EASSE}{Easier Automatic Sentence Simplification Evaluation}

\newacronym{SAMSA}{SAMSA}{Simplification Automatic evaluation Measure through Semantic Annotation}

\newacronym{SARI}{SARI}{System output Against References
and against the Input sentence}

\newacronym{BLEU}{BLEU}{Bilingual Evaluation Understudy Score}

\newacronym{GPT}{GPT}{Generative Pre-trained Transformer}

\usetikzlibrary{matrix}
\pgfplotsset{width=11cm,compat=1.9}
\usepgfplotslibrary{external}
\usepackage{nomencl}
\makenomenclature
\usepackage{etoolbox}

\usepackage{soul,color} 
\usepackage{changepage}

\usepackage{lineno}

\input epsf

\title{A Novel Dataset for Financial Education Text Simplification in Spanish}

\author {\textbf{Nelson Pérez Rojas,$^1$ $^2$} \textbf{Saúl Calderón Ramírez,$^3$} \textbf{Martín Solís Salazar,$^4$} \\ \textbf{Mario Romero Sandoval,$^5$} \textbf{Mónica Arias Monge,$^6$} \textbf{Horacio Saggion$^7$}\\
$^1$Doctorado en Ciencias Naturales para el Desarrollo (DOCINADE), Instituto Tecnológico de Costa Rica,  Costa Rica\\
$^2$Escuela de Ciencias del Lenguaje, Instituto Tecnológico de Costa Rica, Costa Rica\\
$^3$Escuela de Ingeniería en Computación, Instituto Tecnológico de Costa Rica, Costa Rica\\
$^4$Escuela de Administración de Empresas, Instituto Tecnológico de Costa Rica, Costa Rica\\
$^5$Maestría en Computación, Instituto Tecnológico de Costa Rica,  Costa Rica\\
$^6$Instituto de Investigaciones Psicológicas, Universidad de Costa Rica,  Costa Rica\\
$^7$Departamento de Tecnologías de la Información y las Comunicaciones, Universidad Pompeu Fabra,  España\\
nperez@itcr.ac.cr\\
}

\seplntranstitle{Un nuevo conjunto de datos para la simplificación de textos de educación financiera en español}

\seplnresumen{La simplificación de textos es una tarea crucial en el procesamiento del lenguaje natural y consiste en generar textos más comprensibles, especialmente para grupos de lectores específicos. Existen pocos conjuntos de datos en español que se puedan utilizar para crear sistemas de simplificación de textos. En este documento se presenta un conjunto de textos financieros en español destinado a hispanohablantes con discapacidad visual, compuesto por 5,314 pares de oraciones complejas y sus versiones simplificadas elaboradas a partir de reglas. También se expone la comparación del conjunto de datos creado con las simplificaciones generadas por GPT-3, Tuner y MT5, con el fin de evaluar la viabilidad del aumentado de datos utilizando estos modelos. El conjunto de datos está disponible en Hugging Face, como saul1917/FEINA.}

\seplnclave{simplificación de textos, conjunto de datos en español, modelos de lenguaje, características lingüísticas}

\seplnabstract{Text simplification, crucial in natural language processing, aims to make texts more comprehensible, particularly for specific groups like visually impaired Spanish speakers, a less-represented language in this field. In Spanish, there are few datasets that can be used to create text simplification systems. Our research has the primary objective to develop a Spanish financial text simplification dataset. We created a dataset with 5,314 complex and simplified sentence pairs using established simplification rules. We also compared our dataset with the simplifications generated from GPT-3, Tuner, and MT5, in order to evaluate the feasibility of data augmentation using these systems. In this manuscript we present the characteristics of our dataset and the findings of the comparisons with other systems. The dataset is available at Hugging face, as \textit{saul1917/FEINA}.}

\seplnkey{text simplification, datasets in spanish, large language models, linguistic characteristics}

\newpage

\begin{document}

\label{firstpage} \maketitle

%


\section{Introduction} \label{sec:introduction}

Text simplification encompasses a variety of tasks, among them sentence structure modification through different actions such as irrelevant information suppression, splitting long sentences, changing a sentence from a passive form to a subject–verb–object structure, and other actions \cite{saggion2015making,romero2022towards}. Text simplification holds particular value for individuals with visual impairments, as it enables more efficient access to the core meaning of complex texts. Such texts frequently present syntactic and lexical challenges \cite{monge2023analisis}. These complex texts are often syntactically and lexically challenging, particularly in specialized domains like finance. 

One of the main challenges to generating those models is data availability \cite{coster2011simple}. Most existing datasets focus on English, complicating research and development for text simplification in other languages. Most datasets in other languages have been created by aligning sentences from original texts with simpler versions in similar texts \cite{romero2022towards}. The alignment helps to create a more extensive dataset, but it is not always clear which simplification rules were used to make the simple versions. 

Furthermore, creating these datasets needed to adequately consider the unique needs of specific populations, with text simplification requirements varying across different groups \cite{fajardoaportes,romero2022towards}. For example, deaf individuals need simplifications tailored to their experiences with oral language as a second language and the specificities of sign language \cite{monge2023analisis}, differing significantly from the needs of blind individuals who rely on audio formats. Blind people's challenges in accessing information stem from the modalities of obtaining it, not from cognitive limitations. They depend on screen readers to convert text to audio, which requires a different approach to information processing \cite{sauvan2020text,barbieri2005multiabile}. Text simplification, therefore, is not just about comprehension but about making content more accessible auditorily \cite{parmanto2005access}.

In Spanish, limited datasets are available for text simplification, notably the manually curated simplex dataset \cite{saggion2015making} and one developed through automatic sentence alignment from Newsela \cite{palmero2019neural}. This manuscript introduces a novel dataset for Spanish, mainly catering to a specific population, a first of its kind. Our goal is to foster research in text simplification for specialized groups. The focus on financial education aligns with the global movement towards financial literacy, essential for societal financial decision-making and well-being, as recognized in various national strategies \cite{atkinson2013promoting}.

Creating datasets for text simplification, especially in non-English languages and for targeted demographic groups, presents two primary challenges. Firstly, the scarcity of large, high-quality datasets restricts the effectiveness of text simplification models. Secondly, variations in data quality and consistency affect the models' learning efficiency. This manuscript introduces a tailored Spanish financial text simplification dataset to address these issues, consisting of 5,314 sentence pairs developed using distinct simplification rules, thereby enriching the resources available for improving text simplification models.

Moreover, the limited size of manually generated datasets, such as ours, poses a challenge for training Large Language Models (LLMs) that typically require vast datasets \cite{tida2022universal}. Consequently, this study also aims to explore using models like GPT-3, mT5, and Tuner for data augmentation in text simplification. GPT-3, having demonstrated superior results in our evaluations, is considered even more compelling for data augmentation due to its proprietary nature.

The following is a description of the key contributions of this work:

\begin{enumerate}
    \item A set of text simplification rules designed explicitly for Spanish aimed at enhancing the comprehensibility of financial education texts for individuals with visual impairments who rely on screen readers like NVDA and JAWS \cite{singh2021chartsight}.
    \item A novel text simplification dataset in Spanish using the rules defined previously, which we refer to as \gls{FEINA}.
    \item A data-oriented evaluation of the quality and reliability of the manually and automatically generated datasets.

\end{enumerate}

\subsection{Text simplification methods}

Using pre-trained models via transfer learning in NLP is critical for improving task-specific performance. However, this approach has been predominantly focused on English, overlooking many non-English languages. To address such a gap, tmT5 and mC4 are multilingual versions of the T5 model and C4 dataset, respectively. mT5, following the T5's architecture, achieves advanced results in diverse evaluations bolstered by mC4, which includes text in 101 languages \cite{xue2020mt5}.

These models tackle the previously overlooked 'erroneous translation' phenomenon in multilingual NLP models during zero-shot scenarios, where generative models may unintentionally translate text into the wrong language. 

TUNER, a novel model in lexical simplification focusing on Spanish, Portuguese, Catalan, and Galician, utilizes an unsupervised approach to achieve state-of-the-art results, especially in Spanish. Unlike deep learning models like LSBert \cite{qiang2020lsbert}, which rely on dynamic methods, TUNER uses static resources such as vocabularies and thesauri. Its multilingual architecture includes stages like Sentence Analysis, Word Sense Disambiguation (WSD), Synonym Ranking, and Morphological Generation. It also features a morphological generator using lexicon-based algorithms and decision-tree predictions for grammatically accurate inflections \cite{vstajner2022lexical}.

Recent advancements in Text Simplification (TS) systems are driven by large datasets and powerful language models like GPT-3, enhancing their role in NLP applications such as machine translation, text summarization, and question answering. The effectiveness of TS systems hinges on the quality and diversity of training data. In general, to improve this, researchers are exploring strategies like data augmentation, transfer learning, semi-supervised learning and active learning in \gls{NLP} or other domains \cite{ott2019fairseq,calderon2022dealing,benavides2022improving,calderon2022semisupervised,calderon2022dealing,calderon2021correcting,calderon2021dealing,calderon2022real,calderon2022dataset,calderon2021improving,calderon2021improving}.

Developing new AI models, such as OpenAI's ChatGPT, marks a significant advancement. ChatGPT, based on the GPT architecture, utilizes neural networks for natural language management, enabling context-sensitive, human-like interactions. Beyond traditional Text Simplification (TS) uses, ChatGPT offers applications like real-time interaction emulation. However, its adoption comes with challenges, including potential bias in training datasets and the risk of misinformation spread. These issues highlight the importance of careful data curation and ongoing improvement of data enhancement strategies for TS system integrity and accuracy \cite{sallam2023chatgpt}.

Transfer learning in Text Simplification (TS) has become increasingly popular, enhancing TS models by utilizing pre-trained language models (PLMs) like BERT, RoBERTa, and T5 \cite{devlin2018bert}. This method involves fine-tuning a PLM for specific TS tasks, enabling models to recognize complex linguistic features better and improve generalization. Furthermore, a comparative analysis of leading TS datasets for Spanish and English, including ALEXSIS, EASIER/EASIER-500, Simple Wikipedia (SWiRL), Newsela, WikiLarge, and Parallel Universal Dependencies (PUD), focuses on language, sentence count, alignment, domain specificity, and target population \cite{ferres2022alexsis}.

\subsection{Datasets in Text Simplification Research}

Research in automatic text simplification has predominantly been in English. However, efforts have extended to languages like Italian \cite{palmero2019neural}, Japanese \cite{maruyama2020extremely}, French \cite{todirascu2022hector}, German \cite{gonzales2021new}, Spanish \cite{palmero2019neural}, Russian \cite{dmitrieva2021creating}, and Bangla \cite{hossain2021bert}. 
 
These studies have led to the creation of new training and evaluation datasets. Often, these datasets are formed by aligning sentences from original texts with their simplified versions \cite{palmero2019neural,dmitrieva2021creating}, but the exact methods and rules used in simplification are usually kept a secret. In the French context \cite{todirascu2022hector}, a parallel corpus with 4,596 sentence pairs was expanded using a translated version of the WikiLarge dataset. The Japanese study \cite{maruyama2020extremely} utilized a parallel corpus with three simplifications: word or phrase-level simplification, sentence-level simplification, maintaining the original meaning, and document-level simplification.

Notable datasets in English include the Simple Wikipedia (English) and the Turk Corpus. Simple Wikipedia was compiled by aligning sentences from Wikipedia with its simpler version, Simple English Wikipedia, tailored for children, English learners, and non-native speakers who might find standard Wikipedia articles too complex. This alignment used various similarity measures. The Turk Corpus was derived from a subset of the PWKP dataset \cite{xu2016optimizing,zhu2010monolingual}, filtering sentences with more than a 20\% variation in token count between the complex and simplified forms, mainly retaining paraphrase simplifications. They extracted 2,350 sentences from regular Wikipedia and had them simplified by eight Amazon Mechanical Turk workers, resulting in up to 18,800 complex-simple sentence pairs \cite{xu2016optimizing}.

Regarding Spanish, several datasets have been developed:
The Aligned Newsela corpus (Spanish) \cite{palmero2019neural} contains nearly 1,221 professionally annotated documents. Following cleanup and sentence alignment, 55,890 sentence pairs were obtained as a Spanish benchmark, although this dataset is not available for research.
The SIMPLEXT Corpus \cite{saggion2011text} includes 200 news articles manually simplified by experts to aid comprehension for those with learning disabilities.
The EASIER dataset is annotated for easy-to-read language in lexical simplification, and the ALEXSIS dataset \cite{ferres2022alexsis} has 381 instances for lexical simplification tasks. Each ALEXSIS instance includes a sentence, a complex target word, and 25 substitution options.

\begin{table}[h]
 \resizebox{\columnwidth}{!}{%
\centering
\begin{tabular}{c|cccc}
\hline
\textbf{Dataset} & \textbf{N. Instances} & \textbf{Aligned} & \textbf{Specific Domain} & \textbf{Specific Population} \\ \hline
ALEXSIS          & 381                   & Yes               & No                       & No                           \\
EASIER           & 5130                  & Yes               & No                       & No                           \\
EASIER-500       & 500                   & Yes               & No                       & No                           \\
Newsela          & 20000                 & Yes              & No                       & No                           \\
Med-EASi         & 1979                  & Yes               & Medical             & No                           \\ \hline
\end{tabular}}
\captionsetup{font=scriptsize}
    \caption{Examples of datasets.}
    \label{tab:Examples data sets}
\end{table}

This comparison highlights the limited availability of domain-specific and target population-focused data sets for \gls{TS}, particularly for Spanish. Most of these data sets are also aligned. The latter presents a problem for training \gls{TS} models: more custom \gls{TS} data sets are needed for specific domains and populations, emphasizing alignment to speed up model training. The data sets compared in this study will be used as a reference point for creating and evaluating \gls{TS} data sets in the future.

\section{Methodology}\label{ref:sec_methodology}

This work can be split in three methodological steps, following the research questions established in Section \ref{sec:introduction}

\subsection{Financial Education Text Dataset Construction}
The first step comprises the process of building the text simplification dataset. Initially, we selected 4 books about financial education \cite{torres2020manual,gil2008libro,izaguirre2020finanzas,National} to create the dataset. We selected books that let us release to the scientific research community the dataset. The second step was to extract text segments (sentences) from the books. A text segment is defined as follows: It is a piece of text that is separated from the next segment text by the following punctuation symbols: a) dot, b) semicolon, c) Closing question mark, d) Closing exclamation mark. In summary, we split the books into pieces of text that we called text segments. For example, in table II we can see a paragraph decomposed into text segments, using the dot symbol.

\begin{table}[h]
\renewcommand{\arraystretch}{3}
\resizebox{\columnwidth}{!}{%
\begin{tabular}{p{0.32\textwidth}p{0.32\textwidth}}
\hline 
Paragraph  & Segments\tabularnewline
\hline 
\hline 
\multirow{6}{=}{“La tangibilidad es una característica muy esencial de un objeto para considerarlo bien económico. Tangible es por consiguiente un computador, un carro, una casa, zapatos y muchísimas otras mercancías que son consideradas bienes económicos. Por otro lado, para que un producto adquiera la categoría de bien económico, debe ser satisfactor de una necesidad o un deseo”.}
&La tangibilidad es una característica muy esencial de un objeto para considerarlo bien económico.\tabularnewline
\cline{2-2} 
 &Tangible es por consiguiente un computador, un carro, una casa, zapatos y muchísimas otras mercancías que son consideradas bienes económicos.\tabularnewline
 \cline{2-2} 
  &Por otro lado, para que un producto adquiera la categoría de bien económico, debe ser satisfactor de una necesidad o un deseo.\tabularnewline
\hline 
\end{tabular}}
     \captionsetup{font=scriptsize}
    \caption{Examples of text segments extracted from a paragraph.}
    \label{tab:Examples texts segments}
\end{table}

In the third phase of our research, we formulated a set of clear, exemplified guidelines. These guidelines outlined the attributes within a text segment that required simplification and the necessary actions to craft a more concise text version. Our approach to designing these textual simplification guidelines was rigorous. We conducted an in-depth review of existing literature, specifically focusing on the Simplext corpus developed by the Diles Group at the Autonomous University of Madrid \cite{anula2011pautas}. This literary exploration was further enriched by collaborations with specialists in text adaptation for the visually impaired and linguistic experts, ensuring the fine-tuning of simplification rules vital for optimal compatibility with screen readers. 
Table A1 of the Appendix shows an extract from the manual. The second column of the appendix has the name of attributes that should be simplified, and the third column has a summary of the instructions used to solve the complexity generated by the attribute. For example, the first attribute in column two is named “word frequency” and the instruction in the third column is to replace low-frequency words with commonly used words. In summary, these instructions are the guidelines to generate simple text segments from the original text segments. A total of 21 attributes with their instruction were generated. 

 Finally, six advanced philology students were recruited to simplify text segments using a manual, producing 5,314 pairs of complex/straightforward text segments. The most frequent attributes requiring simplification were superfluous words, word length, and complex lexical expressions, as depicted in Figure 1. Less common attributes, like lexical ambiguity and compound subjects, were underrepresented. The dataset, potentially biased towards more prevalent attributes, reflects the domain-specific nature of the source texts. For example, the abstract nature of finance implies less frequent use of visual references compared to longer word lengths typical of financial jargon. The final dataset, including original texts, simplified versions, and identified attributes, is available online.

\begin{figure}[htp]
    \centering
    \includegraphics[width=8cm]{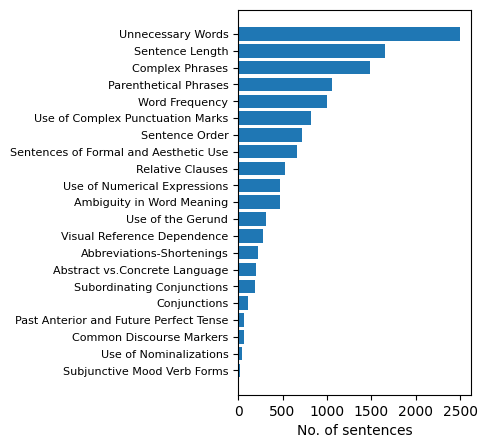}
    \captionsetup{font=scriptsize}
    \caption{Histogram of the the simplification rules used to generate the manually simplified dataset. }
    \label{fig:categories_freq}
\end{figure}

\subsection{Benchmarking against machine-generated simplifications: GPT-3, Tuner and MT5}\

Different methods for text simplification have been developed in the literature, with different approaches;  from rules-based systems to \gls{LLM}. To identify what models are worth to test, we developed a small qualitative test, with the simplification of 13 text segments using the following systems:  \gls{GPT}-3,	MT5,	Tuner,	FastChat,	Alpaca13b and	Llama 13b. A sample of the results can be seen in Table A2 of the Appendix. In most of the examples \gls{GPT}-3  outputs the most aggressive simplifications, shortening the text segments considerably and keeping the main idea of the segments. Llama presents inconsistent text segments, including even words in English, as seen in the first example. Alpaca also presents results with low semantics correctness. Similarly, FastChat fails to deliver semantically correct results. MT5 and Tuner delivered semantically and syntactically correct results. MT5 was able to generate more simple text segments when compared to Tuner, as the former tends to leave the text segment unchanged. According to our evaluation, \gls{GPT}-3 yielded the best results according to correctness, semantic preservation and simplification. Also Tuner and MT5 achieved outputs with considerable correctness, unlike the rest of the models (FastChat, Llama and Alpaca), which often failed in producing syntactically and semantically correct outputs. Other \gls{LLM}s were tested: Bard does not work in Spanish by the writing time of this work, Mosaic MPT-7B model (see \url{https://huggingface.co/spaces/mosaicml/mpt-7b-chat}) did not work well in our preliminary tests. From the previous analysis we decided to select \gls{GPT}-3, MT5 and Tuner for the second step.

As a second step in the implemented methodology of this work, we evaluated \gls{GPT}-3, MT5 and Tuner to perform automatic text simplification using the same source text segments built in the previous. We aim to evaluate the quality of each one of the tested models against the manually generated data. This can shed light on how close automatic systems are to manual simplifications in  a specific domain such as finance education for blind people. Data augmentation policies for example can be formulated upon the measured performance.  

Therefore, we generated the additional text-simplifications datasets
\begin{itemize}
    \item Financial Education Text Dataset simplified by Tuner: built using the Tuner system developed in \cite{ferres2017adaptable}.
    \item Financial Education Text Dataset simplified by MT5: Generated with the MT5 simplification model proposed in \cite{vstajner2022sentence}.   
    \item Financial Education Text Dataset Simplified with \gls{GPT}-3: using the \gls{GPT}-3 model described in \cite{brown2020language}. By the time of writing this work, the model was not available for free usage by OpenAI. We used the following prompt \textit{Simplifique el siguiente texto en español:}. Only one prompt result for each input segment was included in the dataset.
\end{itemize}

\subsection{Quality evaluation of the datasets}\label{subsec:text_simplification}
Three kind of evaluations were generated. First, automatic simplicity metrics were computed for each dataset. As previously mentioned, we categorized these metrics into three classes. The following subset of metrics were selected for each class:
\begin{itemize}
    \item Reference-less metrics for single text segments: We selected specific metrics for Spanish: the Fernandez-Huerta, Gutierrez-Polini and Szigriszt Pazos indices. As complementary measurements, we also report the number of mono and polysyllables, as also the number of syllables and words. The metrics of this category were imported from the \textit{TextStats} package \url{https://pypi.org/project/textstat/} \cite{bansal2021textstat}.
    \item Reference-less simple/complex comparison metrics: As for the previously mentioned metrics which account for a comparison between the simple and complex sentences, we measured the compression ratio, number of sentence splits, the Levenshtein similarity index, the number of exact copies and the additions and deletions performed. We used the \gls{EASSE} python package \cite{alva-manchego-etal-2019-easse} to calculate them. To measure the semantic preservation, we used the cosine distance between the \gls{BETO} embeddings of the simple and complex sentences. \gls{BETO} is a \gls{BERT} architecture pre-trained in a Spanish corpus \cite{CaneteCFP2020}.  For sentence embedding extraction we used the library developed in \cite{reimers-2019-sentence-bert} along with the weights of BETO, made available in \url{espejelomar/sentece-embeddings-BETO} through the HuggingFace application programming interface.   

    \item Reference based metrics: As for the quantification of text simplicity based upon 2 or more human created references, we use the \gls{SARI} and \gls{BLEU} scores. We used the implementations of these metrics provided also in the \gls{EASSE} python package. 
    
    \item As a complementary reference-based measurement, we present metrics of discrimination between the complex and simple text segments to determine how much the simplified sentences differ from the original ones. The discrimination metrics are the average Accuracy and F1 score of Random Forest and  Beto models trained with each dataset to classify the text segments into simple or complex (the complex is the original text segment). A higher Accuracy or F1 score of the classifier means that the text segments simplified are easy to discriminate from the original segments.
    
    We fine-tune the Beto model for the binary classification into simple and complex with each dataset. The Random Forests were also trained for the same tasks with each dataset, using TF-IDF. The training was executed with 10 partitions (cross-validation train/test) that were the same for each dataset. Therefore, the index of the text segments in the training and testing were the same between models. The average accuracy and F1 score of tests were presented as metrics of discrimination between the complex and simple text segments.

    In the case of BETO models, the best values of the hyperparameters learning rate, weight decay, batch size, and the scheduler were selected using weights and biases suite with Bayesian optimization. The training dataset in each partition was divided into two subsets: one subset to compute the weights of the last layer of BETO (75\% of the training dataset, 3986 observations) and the other part (25\% of the training dataset, 1328 observations) to define when the model should stop the training using early stopping with the patience of two of the cross-entropy loss function. In the case of Random Forest models, we selected the best values of the number of trees, maximum number of features, and minimum sample of leaf, using Bayesian optimization and cross-validation of 5 folds with each training dataset.

    \item Finally, to incorporate human judgment of the text segments for each system and the manual simplifications, we solicited feedback from two visually impaired subjects on the simplification quality of N = 100 text segments. Both subjects possess extensive expertise in text simplification tailored for blind or visually impaired audiences. They provided their input after completing an informed consent form, ensuring adherence to the ethical considerations of their participation in this study segment.  Each evaluator was asked to reply the following items:
    \begin{itemize}
        \item The output text segment is simpler than the original?
        \item The output text segment is simpler preserves the meaning of the original?
        \item The output text segment is better understood than the original?
        \item The output text is more suitable than the original for blind or visually impaired people?
        
    \end{itemize}
    The items were answered by the two judges using a Likert scale from 1 to 5, where 1 means total agree with the affirmation of the item, 2 agree, 3 neutral, 4 disagree and 5 totally disagree.  
    For each source text segment,  four simplification proposals were presented  to the judges: the manual proposal, the \gls{GPT}-3, tuner and MT5 outputs. The four text simplification proposals were sorted randomly to neutralize a possible  initial order effect. 

\end{itemize}

\section{Results}
 In Subsection  \ref{subsec:metrics_results} the results for the calculated metrics are depicted. We develop the corresponding analysis in order to shed light about the quality of the automatic simplifications.

\subsection{Simplifications metrics for the manual and automatic simplification datasets}\label{subsec:metrics_results}

The first set of experiment results show the output of computing the different reference-less lexical simplicity metrics, the simple/complex comparison of the text segments with no reference, and the reference based metrics, as described in Section \ref{subsec:text_simplification}. We calculate such measures to quantitatively compare how the simplification was carried out for the different tested systems. We also compute the metrics for the manually simplified dataset as a reference. Therefore we include the results for the Manual, \gls{GPT}-3, MT5 large and the rule-based system tested. Finally the results for the human based judgements of a subset of text segments for each generated dataset are described. 

\subsubsection{Reference-less lexical metrics results} \label{subsubsec:referenceless_lexical_metrics}
Tables \ref{tab:manual_lexical_results}, \ref{tab:mt5_lexical_results}, \ref{tab:gpt3_lexical_results} and \ref{tab:tuner_lexical_results}   describe the yielded results of the selected lexical reference-less metrics. As previously mentioned the Fernandez-Huerta, Szigriszt-Pazos, Gutierrez-Polini metrics with higher values attempt to describe a better simplification. As for the word count, polysyllable count in general, the lower, a better text simplification, with a number of exceptions.  
\begin{table}[h]
    \centering
    \resizebox{\columnwidth}{!}{%
    \npdecimalsign{.}
    \nprounddigits{2}
   \begin{tabular}{lrrrrr}
\toprule
{} &   $\mu_s$ &  $\sigma_s$ &   $\mu_c$ &  $\sigma_c$  \\
\midrule
fernandez\_huerta &  87.13 &    14.88 &  79.09 &    20.16  \\
szigriszt\_pazos  &  83.68 &    15.12 &  75.88 &    20.11  \\
gutierrez\_polini &  39.38 &     6.95 &  37.46 &     8.33  \\
crawford         &   3.41 &     1.32 &   3.72 &     1.20  \\
syllable count   &  37.51 &    22.73 &  46.12 &    27.51 \\
word count       &  22.01 &    13.14 &  27.59 &    16.03 \\
poly count       &   4.32 &     3.27 &   5.11 &     3.82  \\
mono count       &  12.27 &     7.82 &  15.97 &     9.63  \\
\bottomrule
\end{tabular}}
    \caption{Statistics for the Manually simplified dataset for the reference-less with no comparison metrics.}
    \label{tab:manual_lexical_results}
\end{table}

Table \ref{tab:manual_lexical_results} shows the lexical complexity metrics for both the simple and complex sentences from the manually simplified dataset. As for the Fernandez-Huerta, Szigriszt-Pasos and the Gutierrez-Polini metrics, a trend where its values are higher in average for the simple sentences can be seen.  Using a Wilcoxon-Mann-Whitney statistical significance test, all of the aforementioned  metrics present a statistically significant difference between the simple and complex tests. This using a confidence interval of 95\%.  

A similar behavior is described in Table \ref{tab:gpt3_lexical_results} for the \gls{GPT}-3 generated dataset, where all the derived metrics yield a higher value for the simple sentences. Also, statistical significance of such metrics between the average of the simple and complex sentences is achieved. The average difference for the derived metrics between the simple and complex sentences is slightly lower when compared to the results yielded in the manually generated dataset. This suggests that manually simplification is better than the \gls{GPT}-3 generated simplifications. 

The Tuner generated dataset exhibits a weaker lexical simplification efectiveness, as seen in Table \ref{tab:tuner_lexical_results}, where the average difference between the derived metrics for the simple and complex sentences is lower when compared to the \gls{GPT}-3 and Manually generated dataset. 

As for the Mt5 simplified dataset, the trend of higher values for the simplified sentences is also visible as seen in Table \ref{tab:mt5_lexical_results}. However, in this case there is no statistical significance for all of them. This according to the Wilcoxon-Mann-Whitney test carried out.  

The highest improvements on lexical simplicity according to the tested metrics were yielded by the manually and \gls{GPT}-3 generated datasets by considerably larger margin. Both Tuner and MT5 achieved the lowest values, suggesting that both systems are considerably worse at lexical simplification.

\begin{table}[h]

    \centering
    \resizebox{\columnwidth}{!}{%
   \begin{tabular}{lrrrrr}
\toprule
{} &   $\mu_s$ &  $\sigma_s$ &   $\mu_c$ &  $\sigma_c$  \\
\midrule
fernandez\_huerta &  83.44 &    14.20 &  82.40 &    17.65 \\
szigriszt\_pazos  &  80.14 &    14.41 &  79.14 &    17.66  \\
gutierrez\_polini &  38.91 &     6.73 &  38.55 &     7.43 \\
crawford         &   3.71 &     1.15 &   3.61 &     1.17  \\
syllable count   &  39.03 &    13.07 &  48.97 &    27.61  \\
word count       &  23.37 &     7.59 &  29.36 &    16.13  \\
poly count       &   4.26 &     2.33 &   5.41 &     3.86  \\
mono count       &  13.45 &     5.06 &  17.04 &     9.74  \\
\bottomrule
\end{tabular}}
    \caption{Statistics for the mT5 text simplifications reference-less metrics.}
    \label{tab:mt5_lexical_results}
\end{table}

\begin{table}[h]
    \centering
    \resizebox{\columnwidth}{!}{%
    \begin{tabular}{lrrrrr}
        \toprule
        {} &   $\mu_s$ &  $\sigma_s$ &   $\mu_c$ &  $\sigma_c$ \\
        \midrule
        fernandez\_huerta &  86.02 &    16.73 &  79.09 &    20.16 \\
        szigriszt\_pazos  &  82.67 &    16.97 &  75.88 &    20.11  \\
        gutierrez\_polini &  39.55 &     7.66 &  37.46 &     8.33  \\
        crawford         &   3.44 &     1.40 &   3.72 &     1.20  \\
        syllable count   &  37.41 &    23.48 &  46.12 &    27.51  \\
        word count       &  22.22 &    13.57 &  27.59 &    16.03  \\
        poly count       &   4.18 &     3.33 &   5.11 &     3.82  \\
        mono count       &  12.64 &     8.06 &  15.97 &     9.63 \\
        \bottomrule
    \end{tabular}}
    \caption{Statistics for the GPT-3 text simplifications referenceless metrics.}
    \label{tab:gpt3_lexical_results}
\end{table}

\begin{table}[h]
    \centering
    \resizebox{\columnwidth}{!}{%
   \begin{tabular}{lrrrrr}
\toprule
{} &   $\mu_s$ &  $\sigma_s$ &   $\mu_c$ &  $\sigma_c$ \\
\midrule
fernandez\_huerta &  85.80 &    16.17 &  82.13 &    17.52 \\
szigriszt\_pazos  &  82.48 &    16.31 &  78.86 &    17.52  \\
gutierrez\_polini &  39.93 &     7.11 &  38.49 &     7.38  \\
crawford         &   3.44 &     1.20 &   3.64 &     1.14  \\
syllable count   &  50.48 &    28.99 &  49.20 &    27.62 \\
word count       &  30.17 &    16.80 &  29.48 &    16.12  \\
poly count       &   5.52 &     4.04 &   5.44 &     3.87 \\
mono count       &  17.21 &     9.97 &  17.10 &     9.73  \\
\bottomrule
\end{tabular}}

    \caption{Statistics for the Tuner text simplifications reference-less metrics.}
    \label{tab:tuner_lexical_results}
\end{table}

\subsection{Simple-complex comparison based metrics}

Table \ref{tab:easse_comparison_based_metrics_all} shows the simple-complex comparison based metrics calculated for the 4 datasets: the compression ratio, the number of sentence splits, the Levenshtein similarity, the number of exact copies, the additions/deletions proportion, the lexical complexity score and the \gls{BERT} and BETO based similarity scores. For the compression ratio, the Manually and \gls{GPT}-3 generated datasets show a clear lead over the MT5 and Tuner generated datasets. This suggests how the two former systems are not as effective as the manual and \gls{GPT}-3 based simplifications, fro m a lexical simplification perspective. The SBert and SBETO based similarity scores suggest that the semantics of the simplified text segments are fairly similar for the \gls{GPT}-3 and manual simplifications, which advocates for a consistent simplification process of both systems. In the case of the Tuner and MT5 systems, the very high Sbert and SBETO similarities of the source and simplified text segments, along with the higher compression ratio and the number of exact copies suggest for a frequently absent text simplification. This means that  identical simple text segments are generated for the MT5 and Tuner systems, when compared to the source text segment. 

Table \ref{tab:easse_comparison_based_metrics_all} also shows that the \textit{exact copies} metric for the \gls{GPT}-3 and manual simplification systems are 0 for both, meaning that there is always an effective text transformation performed in both cases. The proportion of additions and deletions for \gls{GPT}-3 and the manual outputs also correlates with this trend, as both generate text segments with significantly higher transformations of both types, when compared to the Tuner and MT5 systems. The Levenshtein similarity score strengthens this argument, showing lower scores for the manually and \gls{GPT}-3 when compared to the MT5 and Tuner systems. Finally, according to  the lexical complexity score, also the \gls{GPT}-3 and Manual systems managed to generate less lexically complex text segments, when compared to the Tuner and MT5 systems. A similar trend regarding the \textit{sentence splits} metric is revealed, where the manually and \gls{GPT}-3 systems are able to perform more sentence splits when compared to the Tuner and MT5 systems. 
 
For the 9 tested simple-complex text segment based metrics we carried out a normality and homoscedasticity (equal variances) statistical test, to choose which statistic tests to use to compare these 9 variables across the 4 treatments (manually, MT5, Tuner and \gls{GPT}-3 simplified text segments) used in the experimental setting. We used the pingouin python package to implement the normality and the homoscedasticity tests. According to them, no of the variables for the 4 treatments are normal or fulfill homoscedasticity. Therefore we chose to carry out a Wilcoxon signed test.  

Table \ref{tab:easse_comparison_based_metrics_all} shows the statistical analysis using the aforementioned Wilcoxon test of all the datasets for each of the 9 evaluated metrics. Value pairs within the same row underlined or with the * symbol are not statistically  according to the Wilcoxon signed test. It describes how for every comparison between text simplification systems, there is a statistically significant difference, except in the case when comparing the compression ratio between the manually and the \gls{GPT}-3 generated datasets, as also is the case for the comparison between both systems using the lexical complexity score. Finally, when comparing the  MT5 and the Tuner datasets, the lexical complexity score is also not statistically significantly different.  

In summary, according to Table \ref{tab:easse_comparison_based_metrics_all} we can consider that the manual simplifications are slightly better according to the Levenshtein similarity, \textit{sentence splits} and the number of deletions and additions, when compared to the \gls{GPT}-3 based simplification system. As for the compression ratio and lexical complexity score, the \gls{GPT}-3 system performs very similar when compared to the manual simplifications. Both the manual and \gls{GPT}-3
 text simplifications are considerably better compared to the Tuner and MT5 simplifications.

\begin{table}[h]
    \centering
    \resizebox{\columnwidth}{!}{%
   \begin{tabular}{lrrlr}
\hline
                         & \textbf{Manually Ds.} & \textbf{GPT-3 Ds.} & \textbf{Tuner Ds.} & \multicolumn{1}{l}{\textbf{MT5 Ds.}} \\ \hline
Compression ratio        & \textbf{\underline{0.84$\pm$0.05}}             & \textbf{\underline{0.84$\pm$0.13}}          & 1.02$\pm$0.01          & 0.89$\pm$0.03                            \\
Sentence splits          & \textbf{1.36$\pm$0.49}             & 1.16$\pm$0.59          & 1.26$\pm$0.27          & 0.9$\pm$0.1                              \\
Levenshtein similarity   & \textbf{0.76$\pm$0.03}             & 0.77$\pm$0.03          & 0.94$\pm$0.01          & 0.91$\pm$0.02                            \\
Exact copies             & \textbf{0.00$\pm$0.0}              & 0.00$\pm$0.0           & 0.37$\pm$0.23          & 0.29$\pm$0.21                            \\
Additions proportion     & 0.18$\pm$0.02             & 0.16$\pm$0.03          & 0.07$\pm$0.01          & 0.05$\pm$0.01                            \\
Deletions proportion     & \textbf{0.37$\pm$0.04}             & 0.34$\pm$0.04          & 0.05$\pm$0.01          & 0.17$\pm$0.04                            \\
Lexical complexity score & \textbf{\underline{9.55$\pm$0.58}}             & \textbf{\underline{9.55$\pm$0.59}}          & 9.61$\pm$0.45*          & 9.62$\pm$0.44*                            \\
Sbert cos. sim.          & 0.87$\pm$0.01             & 0.89$\pm$0.01          & 0.97$\pm$0.00          & 0.94$\pm$0.01                            \\
Sbeto cos. sim.          & 0.92$\pm$0.01             & 0.92$\pm$0.01          & 0.98$\pm$0.00          & 0.96$\pm$0.00                            \\ \cline{1-5}
\end{tabular}}
    \captionsetup{font=scriptsize}
    \caption{Results for all the tested systems using the reference-less comparison based metrics. Value pairs in the same row underlined or with * are not statistically different according to the Wilcoxon test, with a $p=0.05$. } 
    \label{tab:easse_comparison_based_metrics_all}
\end{table}

\subsection{Reference based metrics}

As for the reference based metrics \gls{SARI} and \gls{BLEU}, are depicted in Tables \ref{tab:sari_results_all} and \ref{tab:bleu_scores_all}, respectively. 
Specifically for the \gls{BLEU} metric, Table \ref{tab:bleu_scores_all} reveals that the manual simplifications perform best through all the dataset variation cases: using two text segment references (with a sample size of 149 observations) up to five text segment references (with a sample size of 94 observations).  In second place, the \gls{GPT}-3 simplifications are the better ones, using the \gls{SARI} metric according to Table \ref{tab:sari_results_all} for all the tested datasets with 2 to 5 references. The decrease of the \gls{SARI} value when compared to the manual simplifications is around 6 units. In third place, the MT5 simplifications yield a value near to 33, for all the tested dataset samples with different number of references. Similarly, a drop of around 6 units for the \gls{SARI} metric is perceived between the \gls{GPT}-3 and MT5 simplifications. Finally, the simplifications with the lowest \gls{SARI} values in average, were yielded by the Tuner system.  

As for the statistical relevance of the difference between each system according to the \gls{SARI} and \gls{BLEU} metrics, we verified that nor the \gls{BLEU} or \gls{SARI} metric present a normal distribution and equal variances for both the manual and \gls{GPT}-3 systems. Therefore, similar as the previous set of metrics, we executed a non-parametric Wilcoxon signed test to verify whether there is a statistically significant difference between the manual and \gls{GPT}-3 metrics. In the same fashion as the results for the previous metrics set, we marked in Tables \ref{tab:sari_results_all} and \ref{tab:bleu_scores_all} the pairs of statistically similar results (using a $p=0.05$ as criteria). According to the statistical analysis results, only the \gls{BLEU} metric based comparison of the manual and \gls{GPT}-3 systems can be considered not to be statistically different.

\begin{table}[h]
    \centering
    \resizebox{\columnwidth}{!}{%
    \begin{tabular}{ccrrrr}
    
\hline
\textbf{N. r.} & \textbf{N.} & \multicolumn{1}{c}{\textbf{Manual}} & \multicolumn{1}{c}{\textbf{GPT-3}} & \multicolumn{1}{c}{\textbf{MT5}} & \multicolumn{1}{c}{\textbf{Tuner}} \\ \hline
2               & 149            & \textbf{48.7$\pm$12.6}                     & 41.6$\pm$12.2                    & 34$\pm$11.3                  & 26.4$\pm$7.9                     \\
3               & 134            & \textbf{48.2$\pm$11.9}                     & 42.4$\pm$11.7                    & 34.1$\pm$11.4                   & 26.7$\pm$8.1                     \\
4               & 119           & \textbf{48.1$\pm$11.9}                     & 42.1$\pm$11.3                    & 33.9$\pm$11.2                  & 26.7$\pm$8.2                     \\
5               & 94             & \textbf{48.8$\pm$10.7}                     & 41.1$\pm$10                    & 34.4$\pm$11.1                  & 27.4$\pm$8.5                     \\ \hline
\end{tabular}
}
    \captionsetup{font=scriptsize}
    \caption{Average and standard deviation of the \gls{SARI} scores for the simplified text segments generated by the manual, \gls{GPT}-3, Tuner and MT5 systems. The higher the better.}
    \label{tab:sari_results_all}
\end{table}

\begin{table}[h]
    \centering
     \resizebox{\columnwidth}{!}{%
    \begin{tabular}{ccrrrr}
\hline
\textbf{N. r.} & \textbf{N.} & \multicolumn{1}{c}{\textbf{Manual}} & \multicolumn{1}{c}{\textbf{GPT-3}} & \multicolumn{1}{c}{\textbf{MT5}} & \multicolumn{1}{c}{\textbf{Tuner}} \\ \hline
2               & 149              & 49$\pm$23.1                               & 44.9$\pm$22.1                            & \textbf{56.4$\pm$17.8}                          & 51.1$\pm$17.7                            \\
3               & 134              & \underline{48.5$\pm$22.5}                               & \underline{45.5$\pm$22.1}                            & \textbf{56.2$\pm$17.1}                          & 51.4$\pm$17.1                            \\
4              & 119              & \underline{47.5$\pm$22.3}                               & \underline{44.9$\pm$22.1}                            & \textbf{54.9$\pm$17}                          & 50.4$\pm$17.1                            \\
5               & 94               & \underline{47.9$\pm$20.9}                               & \underline{44.81$\pm$21.6}                            & \textbf{55.6$\pm$16}                          & 51$\pm$15.2                            \\ \hline
\end{tabular}
}
    \captionsetup{font=scriptsize}
    \caption{Average and standard deviation of the \gls{BLEU} scores for the simplified text segments generated by the manual, \gls{GPT}-3, Tuner and MT5 systems. The higher the better. Value pairs underlined are no statistically different according to the Wilcoxon test with a $p=0.05$}
    \label{tab:bleu_scores_all}
\end{table}

\subsection{Discrimination between complex and simple versions}

Table \ref{tab:acc_f1_score_classifier_simplicity}shows the  performance metrics of the models trained to classify text segments as complex or simple. Our assumption is that the classifier should yield higher accuracy with better simplifications. Using the Wilcoxon test, we found that the Beto and Random Forest classifiers trained with the manually dataset, have better accuracy and F1 score than the models trained with the other datasets. Therefore, we can conclude that the manual dataset has more clear discrimination between simple and complex sentences. It is important to clarify that we considered the original text segments as complex because they had complex attributes that had to be simplified according to the advanced students in Philology.
 
On the other hand, we found that \gls{GPT}-3 classifiers  tend to get similar performance to turner classifiers, and better performance to MT5 with the exception of the accuracy in \gls{BETO}. Therefore, we conclude that \gls{GPT}-3 and Turner discriminate between complex and simple text segments in a similar way.

\begin{table}[h]
  \centerline{
  \resizebox{\columnwidth}{!}{%
    \begin{tabular}{l|cc|cc}
\hline
\multirow{2}{*}{\textbf{Dataset}} & \multicolumn{2}{c|}{\textbf{BETO}}   & \multicolumn{2}{c}{\textbf{Random Forest}} \\
                                  & \textbf{Acc.}    & \textbf{F1-score} & \textbf{Acc.}       & \textbf{F1-score}    \\ \hline
Manual                         &   \textbf{0.830 $\pm$0.012} & \textbf{0.830$\pm$0.012}   & \textbf{0.681 $\pm$0.011}    & \textbf{0.681$\pm$0.011}      \\
GPT-3                             & \underline{0.753$\pm$0.009}  & 0.750$\pm$0.010   & \underline{0.637$\pm$0.010}     & \underline{0.636$\pm$0.012}      \\
mT5-large                         & \underline{0.724$\pm$0.012}  & 0.697$\pm$0.010   & 0.600$\pm$0.019     & 0.597$\pm$0.018      \\
Tuner                             & \underline{0.744$\pm$0.006}  & 0.765$\pm$0.008   & \underline{0.630$\pm$0.007}     & \underline{0.629$\pm$0.007}      \\ \hline
\end{tabular}
}
}
    \captionsetup{font=scriptsize}
  \caption{10-Folds mean statistics of the accuracy and F1-score for the BETO based classifier and Random Forest model with a TF-IDF based representation. Value pairs underlined are no statistically different according to the Wilcoxon test with a $p=0.05$}
  \label{tab:acc_f1_score_classifier_simplicity}
\end{table}

\subsection{Human evaluation based metrics}

Table \ref{Table_human_scores} presents the human judgment evaluation results, showing that both manual and GPT-3 datasets received similar and optimistic assessments. The first judge's average scores, being less than two or close to it, suggest that the simplified text segments from both datasets generally enhance the original texts in terms of simplicity, understandability, and appropriateness for blind people while retaining the original meaning. The second judge assigned slightly higher scores, still below 3, indicating similar improvements. While both datasets scored comparably, specific differences emerged: the manual dataset was rated higher for preserving original sentence meanings, whereas the GPT-3 dataset excelled in understandability. The other two datasets received lower ratings across all four dimensions; in many instances, the simplified versions were not deemed superior to the originals.

\begin{table}[h]
    \centerline{
     \resizebox{\columnwidth}{!}{%
   \begin{tabular}{lrrlr}

\hline
                         & \textbf{Manually Ds.} & \textbf{GPT-3 Ds.} & \textbf{Tuner Ds.} & \multicolumn{1}{l}{\textbf{MT5 Ds.}} \\ \hline
Judge1  \\
Simplicity  &  \textbf{\underline{1.83$\pm$1.19}}   & \textbf{\underline{1.83$\pm$1.26}} & 4.30$\pm$1.09   & 2.62$\pm$1.81 \\                           
Meaning Pres. & \textbf{\underline{1.46$\pm$1.30}}            & \textbf{\underline{1.67$\pm$1.30}}                          & 3.37$\pm$1.66*  & 3.44$\pm$1.79*\\

Understand. & \textbf{\underline{2.05$\pm$1.30}}              & \textbf{\underline{1.93$\pm$1.22}}  & 4.26$\pm$1.14          & 2.93$\pm$ 1.70                           \\
Suitability     & \textbf{\underline{2.00$\pm$1.35}}             & \textbf{\underline{2.57$\pm$1.48}} & 4.34$\pm$1.12                      & 2.96 $\pm$1.72                            \\
Judge2  \\
Simplicity & \textbf{\underline{2.46$\pm$1.48}}             & \textbf{\underline{2.39$\pm$1.48}}  & 4.55$\pm$0.94          & 3.46$\pm$1.51                          \\
Meaning Pres.   & \textbf{\underline{1.99$\pm$1.44}}             & \textbf{\underline{2.07$\pm$1.54}}          & 4.02$\pm$1.50*   & 4.00$\pm$1.40*                                            \\

Understand.   & \textbf{\underline{2.59$\pm$1.45}}                          & \textbf{\underline{2.57$\pm$1.48}}             & 4.58$\pm$0.87 & 4.01$\pm$1.29                                            \\

Suitability          & \textbf{\underline{2.56$\pm$1.44}}                            & \textbf{\underline{2.53$\pm$1.43}}                        & 4.55$\pm$0.88                        & 4.02$\pm$1.26                                            \\ \cline{1-5}
\end{tabular}}}
\captionsetup{font=scriptsize}
    \caption{Average of human scores based judgements using the Likert scale. The lower the better. Value pairs in the same row underlined or with * are not statistically different according to the Wilcoxon test, with a $p=0.05$.}
    \label{Table_human_scores}
\end{table}

\section{Conclusions}
In this work we approached the problem of text simplification within the financial education domain for blind people in Spanish with a datacentric perspective. The presented dataset, which we refer to as FEINA, is the largest dataset up to date to our knowledge, consisting of simple/complex text segment pairs generated in a completely manual fashion, following a set of simplification rules also described in this work.

We are concerned for the lack of high quality simplification datasets in Spanish for specific domains, in this case for financial education. Simplifying financial education texts in Spanish can increase the readability of this kind of materials for a specific population such as visually disabled people. We adhere to the conception of Artificial Intelligence (AI) as a task creator instead of a mere automator, as argued by Acemoglu et. al. in (Acemoglu and Restrepo, 2020). Using AI to increase inclusivity is an important endeavour which requires high quality and open datasets.

In this work we carried out an extensive datacentric analysis of the tested simplification systems.
Four types of metrics were calculated: reference-less, reference-less pair based, reference and human evaluation based. The manual and GPT-3 generated simplifications were the best according to all the types of metrics used.

The findings indicate that while our manually curated dataset sets a high standard, GPT-3's results are close in quality, with MT5 and Tuner needing to catch up. These insights suggest GPT-3's potential in augmenting simplification datasets with minimal manual adjustments, enhancing financial education accessibility for the visually impaired in Spanish-speaking regions

Within the framework of performance evaluation, it is observed that the reference-based SARI metric suggests a significantly better performance of manual simplifications compared to GPT-3 results. However, our findings also indicate no significant difference between the simplifications generated by GPT-3 and the manual ones for visually impaired subjects. This result invites reflection on the suitability of current metrics for evaluating the effectiveness of text simplification systems, particularly regarding
preserving meaning and lexical simplification.

In summary, this study offers valuable insight and opens new paths for future work in automatic text simplification in Spanish within the domain of financial education for visually impaired individuals. It is a call to action for researchers in artificial intelligence and natural language processing to promote inclusion by developing increasingly effective and accessible systems and approaches.

\bibliographystyle{fullname}
\bibliography{ref}

\begin{thebibliography}{}

\bibitem[\protect\citename{Alva-Manchego \bgroup et al.\egroup }2019]{alva-manchego-etal-2019-easse}
Alva-Manchego, F., L.~Martin, C.~Scarton, and L.~Specia.
\newblock 2019.
\newblock {EASSE}: {E}asier automatic sentence simplification evaluation.
\newblock In {\em Proceedings of the 2019 Conference on Empirical Methods in Natural Language Processing and the 9th International Joint Conference on Natural Language Processing (EMNLP-IJCNLP): System Demonstrations}, pages 49--54, Hong Kong, China, November. Association for Computational Linguistics.

\bibitem[\protect\citename{Anula}2011]{anula2011pautas}
Anula, A.
\newblock 2011.
\newblock Pautas b{\'a}sicas de simplificaci{\'o}n textual y dise{\~n}o del corpus simplext.
\newblock Technical report, Technical report, Grupo DILES. Madrid, Spain: Universidad Aut{\'o}noma de Madrid.

\bibitem[\protect\citename{Atkinson and Messy}2013]{atkinson2013promoting}
Atkinson, A. and F.-A. Messy.
\newblock 2013.
\newblock Promoting financial inclusion through financial education: Oecd/infe evidence, policies and practice.
\newblock {\em OECD Working Papers on Finance, Insurance and Private Pensions No. 34}.

\bibitem[\protect\citename{Bansal and Aggarwal}2021]{bansal2021textstat}
Bansal, S. and C.~Aggarwal.
\newblock 2021.
\newblock textstat.
\newblock {\em Retrieved September 1st}.

\bibitem[\protect\citename{Barbieri \bgroup et al.\egroup }2005]{barbieri2005multiabile}
Barbieri, T., A.~Bianchi, L.~Sbattella, F.~Carella, M.~Ferra, et~al.
\newblock 2005.
\newblock Multiabile: A multimodal learning environment for the inclusion of impaired e-learners using tactile feedbacks, voice, gesturing, and text simplification.
\newblock {\em Assist Technol: From Virtuality to Real}, 16(1):406--10.

\bibitem[\protect\citename{Benavides-Mata and Calderon-Ramirez}2022]{benavides2022improving}
Benavides-Mata, I. and S.~Calderon-Ramirez.
\newblock 2022.
\newblock Improving semi-supervised deep learning by using automatic thresholding to deal with out of distribution data for covid-19 detection using chest x-ray images.
\newblock {\em arXiv preprint arXiv:2211.02142}.

\bibitem[\protect\citename{Brown \bgroup et al.\egroup }2020]{brown2020language}
Brown, T., B.~Mann, N.~Ryder, M.~Subbiah, J.~D. Kaplan, P.~Dhariwal, A.~Neelakantan, P.~Shyam, G.~Sastry, A.~Askell, et~al.
\newblock 2020.
\newblock Language models are few-shot learners.
\newblock {\em Advances in neural information processing systems}, 33:1877--1901.

\bibitem[\protect\citename{Calderon-Ramirez \bgroup et al.\egroup }2021a]{calderon2021dealing}
Calderon-Ramirez, S., R.~Giri, S.~Yang, A.~Moemeni, M.~Umana, D.~Elizondo, J.~Torrents-Barrena, and M.~A. Molina-Cabello.
\newblock 2021a.
\newblock Dealing with scarce labelled data: Semi-supervised deep learning with mix match for covid-19 detection using chest x-ray images.
\newblock In {\em 2020 25th International Conference on Pattern Recognition (ICPR)}, pages 5294--5301. IEEE.

\bibitem[\protect\citename{Calderon-Ramirez \bgroup et al.\egroup }2021b]{calderon2021improving}
Calderon-Ramirez, S., D.~Murillo-Hernandez, K.~Rojas-Salazar, L.-A. Calvo-Valverd, S.~Yang, A.~Moemeni, D.~Elizondo, E.~L{\'o}pez-Rubio, and M.~A. Molina-Cabello.
\newblock 2021b.
\newblock Improving uncertainty estimations for mammogram classification using semi-supervised learning.
\newblock In {\em 2021 International Joint Conference on Neural Networks (IJCNN)}, pages 1--8. IEEE.

\bibitem[\protect\citename{Calderon-Ramirez \bgroup et al.\egroup }2022a]{calderon2022real}
Calderon-Ramirez, S., D.~Murillo-Hernandez, K.~Rojas-Salazar, D.~Elizondo, S.~Yang, A.~Moemeni, and M.~Molina-Cabello.
\newblock 2022a.
\newblock A real use case of semi-supervised learning for mammogram classification in a local clinic of costa rica.
\newblock {\em Medical \& biological engineering \& computing}, 60(4):1159--1175.

\bibitem[\protect\citename{Calderon-Ramirez \bgroup et al.\egroup }2022b]{calderon2022dataset}
Calderon-Ramirez, S., L.~Oala, J.~Torrents-Barrena, S.~Yang, D.~Elizondo, A.~Moemeni, S.~Colreavy-Donnelly, W.~Samek, M.~A. Molina-Cabello, and E.~Lopez-Rubio.
\newblock 2022b.
\newblock Dataset similarity to assess semisupervised learning under distribution mismatch between the labeled and unlabeled datasets.
\newblock {\em IEEE Transactions on Artificial Intelligence}, 4(2):282--291.

\bibitem[\protect\citename{Calderon-Ramirez, Yang, and Elizondo}2022]{calderon2022semisupervised}
Calderon-Ramirez, S., S.~Yang, and D.~Elizondo.
\newblock 2022.
\newblock Semisupervised deep learning for image classification with distribution mismatch: A survey.
\newblock {\em IEEE Transactions on Artificial Intelligence}, 3(6):1015--1029.

\bibitem[\protect\citename{Calderon-Ramirez \bgroup et al.\egroup }2022]{calderon2022dealing}
Calderon-Ramirez, S., S.~Yang, D.~Elizondo, and A.~Moemeni.
\newblock 2022.
\newblock Dealing with distribution mismatch in semi-supervised deep learning for covid-19 detection using chest x-ray images: A novel approach using feature densities.
\newblock {\em Applied Soft Computing}, 123:108983.

\bibitem[\protect\citename{Calderon-Ramirez \bgroup et al.\egroup }2021]{calderon2021correcting}
Calderon-Ramirez, S., S.~Yang, A.~Moemeni, D.~Elizondo, S.~Colreavy-Donnelly, L.~F. Chavarria-Estrada, and M.~A. Molina-Cabello.
\newblock 2021.
\newblock Correcting data imbalance for semi-supervised covid-19 detection using x-ray chest images.
\newblock {\em Applied Soft Computing}, 111:107692.

\bibitem[\protect\citename{Cañete \bgroup et al.\egroup }2020]{CaneteCFP2020}
Cañete, J., G.~Chaperon, R.~Fuentes, J.-H. Ho, H.~Kang, and J.~Pérez.
\newblock 2020.
\newblock Spanish pre-trained bert model and evaluation data.
\newblock In {\em PML4DC at ICLR 2020}.

\bibitem[\protect\citename{Coster and Kauchak}2011]{coster2011simple}
Coster, W. and D.~Kauchak.
\newblock 2011.
\newblock Simple english wikipedia: a new text simplification task.
\newblock In {\em Proceedings of the 49th Annual Meeting of the Association for Computational Linguistics: Human Language Technologies}, pages 665--669.

\bibitem[\protect\citename{Devlin \bgroup et al.\egroup }2018]{devlin2018bert}
Devlin, J., M.-W. Chang, K.~Lee, and K.~Toutanova.
\newblock 2018.
\newblock Bert: Pre-training of deep bidirectional transformers for language understanding.
\newblock {\em arXiv preprint arXiv:1810.04805}.

\bibitem[\protect\citename{Dmitrieva and Tiedemann}2021]{dmitrieva2021creating}
Dmitrieva, A. and J.~Tiedemann.
\newblock 2021.
\newblock Creating an aligned russian text simplification dataset from language learner data.
\newblock In {\em Proceedings of the 8th Workshop on Balto-Slavic Natural Language Processing}. ACL Anthology.

\bibitem[\protect\citename{Fajardo \bgroup et al.\egroup }2015]{fajardoaportes}
Fajardo, I., A.~Ferrer, V.~{\'A}vila, L.~Gil, L.~Salmer{\'o}n, and M.~G{\'o}mez.
\newblock 2015.
\newblock Aportes de la psicoling{\"u}{\'\i}stica a la simplificaci{\'o}n de textos contributions from psycholinguistics to text simplification.
\newblock {\em IX JORNADAS CIENTÍFICAS INTERNACIONALES DE INVESTIGACIÓN SOBRE PERSONAS CON DISCAPACIDAD}.

\bibitem[\protect\citename{Ferr{\'e}s and Saggion}2022]{ferres2022alexsis}
Ferr{\'e}s, D. and H.~Saggion.
\newblock 2022.
\newblock Alexsis: a dataset for lexical simplification in spanish.
\newblock In {\em Proceedings of the Thirteenth Language Resources and Evaluation Conference}, pages 3582--3594.

\bibitem[\protect\citename{Ferr{\'e}s, Saggion, and G{\'o}mez~Guinovart}2017]{ferres2017adaptable}
Ferr{\'e}s, D., H.~Saggion, and X.~G{\'o}mez~Guinovart.
\newblock 2017.
\newblock An adaptable lexical simplification architecture for major {I}bero-{R}omance languages.
\newblock In E.~Bender, H.~Daum{\'e}~III, A.~Ettinger, and S.~Rao, editors, {\em Proceedings of the First Workshop on Building Linguistically Generalizable {NLP} Systems}, pages 40--47, Copenhagen, Denmark, September. Association for Computational Linguistics.

\bibitem[\protect\citename{for Financial~Education}2017]{National}
for Financial~Education, N.~E.
\newblock 2017.
\newblock {\em Tus gastos, tus ahorros, tu futuro: Guía de preparación financiera para principiantes}.
\newblock National Endowment for Financial Education.

\bibitem[\protect\citename{Gil}2008]{gil2008libro}
Gil, G.
\newblock 2008.
\newblock Libro maestro de educaci{\'o}n financiera. sistema para vivir mejor.

\bibitem[\protect\citename{Gonzales \bgroup et al.\egroup }2021]{gonzales2021new}
Gonzales, A.~R., N.~Spring, T.~Kew, M.~Kostrzewa, A.~S{\"a}uberli, M.~M{\"u}ller, and S.~Ebling.
\newblock 2021.
\newblock A new dataset and efficient baselines for document-level text simplification in german.
\newblock In {\em Proceedings of the Third Workshop on New Frontiers in Summarization}, pages 152--161.

\bibitem[\protect\citename{Hossain and Ahnaf}2021]{hossain2021bert}
Hossain, N. and A.~Ahnaf.
\newblock 2021.
\newblock Bert-based text simplification approach to reduce linguistic complexity of bangla language.
\newblock In {\em 2021 International Conference on Intelligent Technology, System and Service for Internet of Everything (ITSS-IoE)}, pages 1--5. IEEE.

\bibitem[\protect\citename{Izaguirre~Olmedo, Carhuancho~Mendoza, and Silva~Siu}2020]{izaguirre2020finanzas}
Izaguirre~Olmedo, J., I.~M. Carhuancho~Mendoza, and D.~Silva~Siu.
\newblock 2020.
\newblock {\em Finanzas para no financieros}.
\newblock GUAYAQUIL/UIDE/2020.

\bibitem[\protect\citename{Maruyama and Yamamoto}2020]{maruyama2020extremely}
Maruyama, T. and K.~Yamamoto.
\newblock 2020.
\newblock Extremely low-resource text simplification with pre-trained transformer language model.
\newblock {\em International Journal of Asian Language Processing}, 30(01):2050001.

\bibitem[\protect\citename{Monge, Alvarado, and Rojas}2023]{monge2023analisis}
Monge, M.~A., L.~S. Alvarado, and G.~R. Rojas.
\newblock 2023.
\newblock An{\'a}lisis de idoneidad de un banco de {\'\i}tems para personas con discapacidad auditiva y visual en una prueba estandarizada de acceso a la educaci{\'o}n superior en costa rica.
\newblock {\em Actualidades Investigativas en Educaci{\'o}n}, 23(2):1--35.

\bibitem[\protect\citename{Ott \bgroup et al.\egroup }2019]{ott2019fairseq}
Ott, M., S.~Edunov, A.~Baevski, A.~Fan, S.~Gross, N.~Ng, D.~Grangier, and M.~Auli.
\newblock 2019.
\newblock fairseq: A fast, extensible toolkit for sequence modeling.
\newblock {\em arXiv preprint arXiv:1904.01038}.

\bibitem[\protect\citename{Palmero~Aprosio \bgroup et al.\egroup }2019]{palmero2019neural}
Palmero~Aprosio, A., S.~Tonelli, M.~Turchi, M.~Negri, and A.~Di~Gangi~Mattia.
\newblock 2019.
\newblock Neural text simplification in low-resource conditions using weak supervision.
\newblock In {\em Proceedings of the Workshop on Methods for Optimizing and Evaluating Neural Language Generation (NeuralGen)}, pages 37--44. Association for Computational Linguistics (ACL).

\bibitem[\protect\citename{Parmanto \bgroup et al.\egroup }2005]{parmanto2005access}
Parmanto, B., R.~Ferrydiansyah, A.~Saptono, L.~Song, I.~W. Sugiantara, and S.~Hackett.
\newblock 2005.
\newblock Access: accessibility through simplification \& summarization.
\newblock In {\em Proceedings of the 2005 international cross-disciplinary workshop on web accessibility (W4A)}, pages 18--25.

\bibitem[\protect\citename{Qiang \bgroup et al.\egroup }2020]{qiang2020lsbert}
Qiang, J., Y.~Li, Y.~Zhu, Y.~Yuan, and X.~Wu.
\newblock 2020.
\newblock Lsbert: a simple framework for lexical simplification.
\newblock {\em arXiv preprint arXiv:2006.14939}.

\bibitem[\protect\citename{Reimers and Gurevych}2019]{reimers-2019-sentence-bert}
Reimers, N. and I.~Gurevych.
\newblock 2019.
\newblock Sentence-bert: Sentence embeddings using siamese bert-networks.
\newblock In {\em Proceedings of the 2019 Conference on Empirical Methods in Natural Language Processing}. Association for Computational Linguistics, 11.

\bibitem[\protect\citename{Romero \bgroup et al.\egroup }2022]{romero2022towards}
Romero, M., S.~Calder{\'o}n-Ram{\'\i}rez, M.~Sol{\'\i}s, N.~P{\'e}rez-Rojas, M.~Chac{\'o}n-Rivas, and H.~Saggion.
\newblock 2022.
\newblock Towards text simplification in spanish: A brief overview of deep learning approaches for text simplification.
\newblock In {\em 2022 IEEE 4th International Conference on BioInspired Processing (BIP)}, pages 1--7. IEEE.

\bibitem[\protect\citename{Saggion \bgroup et al.\egroup }2011]{saggion2011text}
Saggion, H., E.~G{\'o}mez-Mart{\'\i}nez, E.~Etayo, A.~Anula, and L.~Bourg.
\newblock 2011.
\newblock Text simplification in simplext: Making texts more accessible.
\newblock {\em Procesamiento del lenguaje natural}, (47):341--342.

\bibitem[\protect\citename{Saggion \bgroup et al.\egroup }2015]{saggion2015making}
Saggion, H., S.~{\v{S}}tajner, S.~Bott, S.~Mille, L.~Rello, and B.~Drndarevic.
\newblock 2015.
\newblock Making it simplext: Implementation and evaluation of a text simplification system for spanish.
\newblock {\em ACM Transactions on Accessible Computing (TACCESS)}, 6(4):1--36.

\bibitem[\protect\citename{Sallam}2023]{sallam2023chatgpt}
Sallam, M.
\newblock 2023.
\newblock The utility of chatgpt as an example of large language models in healthcare education, research and practice: Systematic review on the future perspectives and potential limitations.
\newblock {\em medRxiv}.

\bibitem[\protect\citename{Sauvan \bgroup et al.\egroup }2020]{sauvan2020text}
Sauvan, L., N.~Stolowy, C.~Aguilar, T.~Fran{\c{c}}ois, N.~Gala, F.~Matonti, E.~Castet, and A.~Calabr{\`e}se.
\newblock 2020.
\newblock Text simplification to help individuals with low vision read more fluently.
\newblock In N.~Gala and R.~Wilkens, editors, {\em Proceedings of the 1st Workshop on Tools and Resources to Empower People with REAding DIfficulties (READI)}, pages 27--32, Marseille, France, May. European Language Resources Association.

\bibitem[\protect\citename{Singh and Goyal}2021]{singh2021chartsight}
Singh, M. and P.~Goyal.
\newblock 2021.
\newblock Chartsight: An automated scheme for assisting visually impaired in understanding scientific charts.
\newblock In {\em VISIGRAPP (5: VISAPP)}, pages 309--318.

\bibitem[\protect\citename{{\v{S}}tajner \bgroup et al.\egroup }2022]{vstajner2022lexical}
{\v{S}}tajner, S., D.~Ferr{\'e}s, M.~Shardlow, K.~North, M.~Zampieri, and H.~Saggion.
\newblock 2022.
\newblock Lexical simplification benchmarks for english, portuguese, and spanish.
\newblock {\em Frontiers in Artificial Intelligence}, 5:991242.

\bibitem[\protect\citename{{\v{S}}tajner, Sheang, and Saggion}2022]{vstajner2022sentence}
{\v{S}}tajner, S., K.~C. Sheang, and H.~Saggion.
\newblock 2022.
\newblock Sentence simplification capabilities of transfer-based models.
\newblock In {\em Proceedings of the AAAI Conference on Artificial Intelligence}, volume~36, pages 12172--12180.

\bibitem[\protect\citename{Tida, Srinivas, and Hsu}2022]{tida2022universal}
Tida, V.~Srinivas, and S.~Hsu.
\newblock 2022.
\newblock Universal spam detection using transfer learning of bert model.
\newblock {\em arXiv preprint arXiv:2202.03480}.

\bibitem[\protect\citename{Todirascu \bgroup et al.\egroup }2022]{todirascu2022hector}
Todirascu, A., R.~Wilkens, E.~Rolin, T.~Fran{\c{c}}ois, D.~Bernhard, and N.~Gala.
\newblock 2022.
\newblock Hector: A hybrid text simplification tool for raw texts in french.
\newblock In {\em 12th International Conference on Language Resources and Evaluation (LREC)}.

\bibitem[\protect\citename{Torres~Salazar and Ramos~Arriagada}2020]{torres2020manual}
Torres~Salazar, G. and R.~Ramos~Arriagada.
\newblock 2020.
\newblock Manual de finanzas personales y de familia. c{\'o}mo usar bien mi dinero y el tuyo.

\bibitem[\protect\citename{Xu \bgroup et al.\egroup }2016]{xu2016optimizing}
Xu, W., C.~Napoles, E.~Pavlick, Q.~Chen, and C.~Callison-Burch.
\newblock 2016.
\newblock Optimizing statistical machine translation for text simplification.
\newblock {\em Transactions of the Association for Computational Linguistics}, 4:401--415.

\bibitem[\protect\citename{Xue \bgroup et al.\egroup }2020]{xue2020mt5}
Xue, L., N.~Constant, A.~Roberts, M.~Kale, R.~Al-Rfou, A.~Siddhant, A.~Barua, and C.~Raffel.
\newblock 2020.
\newblock mt5: A massively multilingual pre-trained text-to-text transformer.
\newblock {\em arXiv preprint arXiv:2010.11934}.

\bibitem[\protect\citename{Zhu, Bernhard, and Gurevych}2010]{zhu2010monolingual}
Zhu, Z., D.~Bernhard, and I.~Gurevych.
\newblock 2010.
\newblock A monolingual tree-based translation model for sentence simplification.
\newblock In {\em Proceedings of the 23rd International Conference on Computational Linguistics (Coling 2010)}, pages 1353--1361.

\end{thebibliography}

\clearpage
\onecolumn

\appendix
\section{Appendix}

\setcounter{table}{0}
\renewcommand{\thetable}{A\arabic{table}}
\begin{table}[h]
    \centering
     \resizebox{0.9\columnwidth}{!}{%
\begin{tabular}
{p{0.1\textwidth}p{0.3\textwidth}p{0.8\textwidth}}

\textbf{Original} &
  \textbf{Attribute} &
  \textbf{Guideline for simplification}\\ \hline
  
  1 & Word Frequency  & Replace low-frequency words with commonly used words.   \\ \hline
2 &  Abstract vs. Concrete Language & When possible, replace words with abstract or less imaginable referents with concrete or imaginable referents. It can be suppressed if the word is in an unnecessary idea to convey the message. \\ \hline
3 &	Visual Reference Dependence &	When possible, replace discourse segments with visual references. It can be suppressed if the word is in an unnecessary idea to convey the message. \\ \hline
4 &	Ambiguity in Word Meaning &	Replace polysemic words according to their main meanings. \\ \hline
5 &	Unnecessary Words &	Eliminate words and phrases that are not necessary for understanding the message of the text. \\ \hline
6 &	Complex Phrases & Replace phrases (complex lexical expressions) with their one-word equivalents. \\ \hline
7 &	Use of Nominalizations &	Replace nominalizations with verbs that express the same states, processes, and actions. \\ \hline
8 & Conjunctions & \underline{Coordinating}: Preferably use the conjunctions "y" and "ni". However, replacement is unnecessary in short sentences and those that do not employ other attributes that make the text difficult. \underline{Disjunctive} : Use the conjunction "o". However, modification is only recommended when the marker is infrequently used or when a less explanatory style is used for a financial education text. \underline{Adversative} : Use the conjunction "pero" in affirmative constructions and "sino" in negative constructions. However, modification is only recommended when the marker is infrequently used or when a less explanatory style is used for a financial education text. \underline{Other Coordination Connectors}: Modify these coordination links only when the marker is infrequently used or when a less explanatory style is used for a financial education text.\\ \hline 
9  &  Subordinating Conjunctions & \underline{Relative}:
Use the simple relative forms "que", "quien", and "cuanto". Replace the simple form "cuyo". \underline{Temporal}: Preferably use "cuando" and "mientras". \underline{Locative}: Use "donde". \underline{Modal}: Use "como" and "según". \underline{Conditional}: Use "si" and "por si", provided important logical expressions for financial texts are unaffected. \underline{Concessive}: Use "aunque", or other frequently used ones such as "a pesar de que" and "incluso cuando". \\ \hline
10 & Past anterior and future perfect tense & Use the forms of present, simple past, imperfect past, simple future, simple conditional, present perfect, pluperfect, and conditional perfect. \\ \hline
11 & Subjunctive Mood Verb Forms  & Limit the use of subjunctive to its forms of present, imperfect past, simple future, and pluperfect. \\ \hline
12 & Use of Gerund & Replace the gerund forms in the sentences. \\ \hline
13 & Abbreviations, Shortenings, Acronyms, and Initialisms &
 Use full expressions instead of abbreviations, shortenings, acronyms, and initialisms. \\ \hline
14 & Use of Numerical Expressions & Write numerical expressions with letters. \\ \hline
15 & Sentence Length & Use segments of 20 words or less. Dividing the discourse segment into several segments is recommended. \\ \hline
16 & Parenthetical Phrases & Reserve use when they are essential for understanding the message. \\ \hline
17 & Relative Clauses & Reduce the number of relative clauses per sentence (maximum one). \\ \hline
18 & Sentences of formal and aesthetic use &  Modify the use of these constructions when they are used solely for stylistic or aesthetic reasons. \\ \hline
19 & Sentence Order & Preferably use the SVO order. \\ \hline
20 & Common Discourse Markers & Use the most frequent markers in the cultural context. \\ \hline
21 & Use of Complex Punctuation Marks & Eliminate using semicolons, brackets, dashes, single quotes, and angle quotes. Do not change the semicolon to period if the linguistic units separated with the first punctuation mark are words or phrases (enumerating lexical elements). \\ \hline

\end{tabular}
}
    \captionsetup{font=scriptsize}
    \caption{Attributes and Guidelines for Simplifying Discourse Segments of Financial Texts for Visually Impaired Individuals }
    \label{tab:Appendix2}
\end{table}

\setcounter{table}{1}

\begin{table}[h]
    \centering
    \resizebox{1\columnwidth}{!}{%
\begin{tabular}{p{0.2\textwidth}p{0.2\textwidth}p{0.2\textwidth}p{0.2\textwidth}p{0.2\textwidth}p{0.2\textwidth}p{0.2\textwidth}}
\textbf{Original} &
  \textbf{GPT-3} &
  \textbf{MT5} &
  \textbf{Tuner} &
  \textbf{FastChat} &
  \textbf{Alpaca13b} &
  \textbf{Llama 13b} \\ \hline
Cuando se intenta definir el dinero, las personas piensan en monedas o billetes, lo cual es muy reducido porque el dinero es más que eso. &
  El dinero es más que monedas o billetes. &
  Cuando se intenta definir el dinero,  las personas piensan en monedas o billetes,  lo cual es muy reducido. &
  Cuando se intenta definir el dinero,  las personas piensan en monedas o billetes,  lo cual es muy reducido porque el dinero es mas que eso. &
  Cuando se trata de definir el dinero, las personas pensan en monedas o billetes, pero el dinero es más que esos. No se trata sólo de monedas o billetes. &
  El dinero es un concepto más amplio que simplemente monedas o billetes. Se define como una mediodinero es más que eso. Se define como una mediodinero es más que eso. &
  The dinero es una forma de intercambiar bienes y servicios. Todos los bienes y servicios tienen un precio y se pueden comprar y vender. Algunos ejemplos de bienes son los hogares, los automóviles, los muebles de la casa, los alimentos y los cosméticos. Algunos ejemplos de servicios son la limpieza, la cocina, los negocios, la atención médica y la música. \\
  \hline
A fin de mes, podrías tener unos \$50 para depositar en tu cuenta de ahorro, que suman \$600 al año (sin incluir los intereses, si los depositaras en una cuenta que pague intereses). &
  A fin de mes, podrás depositar \$50 en tu cuenta de ahorro, lo que sumará \$600 al año. &
  A fin de mes,  podrí­as tener unos 50 para depositar en tu cuenta de ahorro,  que suman \$600 al año (sin incluir los intereses,  si los depositaras en una cuenta que pague intereses). &
  A fin de mes,  podrías tener unos \$50 para depositar en tu cuenta de ahorro. Esta tu cuenta de ahorro suman \$600 al año (sin admitir los intereses,  si los depositases en una cuenta). Esta una cuenta pague intereses. &
  A finales de mes, puedes tener \$50 para depositar en tu cuenta de ahorro, que puede agregarse a \$600 al año (sin incluir los intereses, si los encuentres en una cuenta que pague intereses). &
  Puedes tener unos \$50 para depositar en tu cuenta de ahorro, que suman \$600 al año (sin incluir los intereses, si los depositaras en una cuenta que pague intereses). &
  Fin de mes podrías tener unos \$50 para depositar en tu cuenta de ahorro, que suman \$600 al año sin incluir los intereses, si los depositaras en una cuenta que pague intereses.

  \\
  \hline

  Ingresos de tipo residual: Son ingresos producto de rendimientos financieros, arriendos de bienes raíces, dividendos, acciones de empresas; es decir, las personas no tienen que realizar un trabajo para recibir su ingreso. & 
  Ingresos residuales: se generan a partir de los rendimientos financieros, los arrendamientos de inmuebles, los dividendos y las acciones de las empresas; es decir, las personas no tienen que realizar un trabajo para obtener este ingreso.
  &
Ingresos de tipo residual: Son ingresos producto de rendimientos financieros,  arriendos de bienes raíces,  dividos,  acciones de empresas; es decir,  las personas no tienen que realizar ese tipo de transacciones.

&
Ingresos de tipo residual: Son ingresos producto de rendimientos financieros,  arriendos de bienes raíces,  dividendos,  acciones de empresas; es decir,  las personas no tienen que realizar un trabajo para recibir su ingreso.

&
Residuos residuales: Son medios de ingreso que no se llevarán en acciones o dividendos, y son hechos a causa de rendimientos financieros. Las personas no tienen que hacer trabajo para recibir estos medios.
&
Ingresos de tipo residual: Son ingresos producto de rendimientos financieros, arriendos de bienes raíces, dividendos, acciones de empresas; es decir, las personas no tienen que realizar un trabajo para recibir su ingreso.
&
Ingresos de tipo residual: Son ingresos producto de rendimientos financieros, arriendos de bienes raíces, dividendos, acciones de empresas; es decir, las personas no tienen que realizar un trabajo para recibir su ingreso.
  
\end{tabular}}
    \captionsetup{font=scriptsize}
    \caption{Comparison of the Spanish text simplification datasets.}
    \label{tab:Appendix1}
\end{table}

\end{document}